\documentclass[10pt,twocolumn,letterpaper]{article}

\usepackage{cvpr}
\usepackage{times}
\usepackage{epsfig}
\usepackage{graphicx}
\usepackage{amsmath}
\usepackage{amssymb}
\usepackage{amsthm}

\usepackage{amsthm}
\usepackage{multirow}
\usepackage{bm}

\newtheorem{thm}{Theorem}
\newtheorem{defn}{Definition}

\usepackage[pagebackref=true,breaklinks=true,letterpaper=true,colorlinks,bookmarks=false]{hyperref}

\cvprfinalcopy 

\def\cvprPaperID{3754} 

\ifcvprfinal\fi
\begin{document}

\title{Training Neural Networks by Using Power Linear Units (PoLUs)}

\author{Yikang Li\thanks{Indicates equal contributions.}, \
Pak Lun Kevin Ding\footnotemark[1], \
Baoxin Li\\
School of Computing, Informatics, and Decision Systems Engineering\\
Arizona State University\\
{\tt\small \{yikangli, kevinding, baoxin.li\}@asu.edu}
}

\maketitle
\begin{abstract}
In this paper, we introduce "Power Linear Unit" (PoLU) which increases the nonlinearity capacity of a neural network and thus helps improving its performance.
PoLU adopts several advantages of previously proposed activation functions.
First, the output of PoLU for positive inputs is designed to be identity to avoid the gradient vanishing problem. 
Second, PoLU has a non-zero output for negative inputs such that the output mean of the units is close to zero, hence reducing the bias shift effect.
Thirdly, there is a saturation on the negative part of PoLU, which makes it more noise-robust for negative inputs.
Furthermore, we prove that PoLU is able to map more portions of every layer's input to the same space by using the power function and thus increases the number of response regions of the neural network.
We use image classification for comparing our proposed activation function with others.
In the experiments, MNIST, CIFAR-10, CIFAR-100,
Street View House Numbers (SVHN) and ImageNet are used as benchmark datasets.
The neural networks we implemented include widely-used ELU-Network, ResNet-50, and VGG16, plus a couple of shallow networks. 
Experimental results show that our proposed activation function outperforms other state-of-the-art models with most networks.
\end{abstract}

\section{Introduction}
In recent years, 
neural networks with many hidden layers, 
which are usually called deep neural networks, 
have been used in a lot of machine learning and visual computing tasks, 
delivering unprecedented good results \cite{alexnet,mltask1,mltask2}. 
Unlike other machine learning approaches relying on linear models (possibly with kernel extensions),
neural networks are inherently nonlinear models because of the nonlinear activation functions used in the neurons. 
Properties and applications of neural networks have been studied \cite{nnbook}. 
Some earlier, 
theoretical results, 
such as the  {\em Universal Approximation Theorem}, 
which states that, 
any continuous function on a compact subset of $\mathbb{R}^n$ can be approximated by a feed-forward network with a single hidden layer containing a finite number of neurons \cite{uniappthm,uniappthm2}, 
have alluded to the potential of deep neural networks, 
although much more and better understanding has yet to be developed.

Recent years have witnessed various efforts to this end. 
In particular, 
there are studies aiming at investigating the reason why deeper networks generally perform better.
It was shown that \cite{numreg0}, 
there exist families of functions which can be approximated much more efficiently by a deep neural network than a shallow one, 
when the number of hidden units in the networks is kept the same. 
In \cite{numreg1,numreg2}, 
researchers investigated the relationship between the the depth of the neural networks and the complexity of functions that are computable by the networks.
It was also shown that, 
when using Rectified Linear Unit (ReLU) \cite{relu} as the activation function, 
the number of response regions of the neural network grows exponentially in the number of hidden layers but in polynomial in the number of neurons in one layer. 
This also appears to be intuitive, 
as the more number of hidden layers a network has, 
the more number of hierarchical nonlinear mappings the inputs will go through, 
hence enabling a more flexible representation by the network. 
One thread of research towards increasing the flexibility of the networks is by using a potentially better activation function, 
which is the source of the nonlinearity of the model. 
Despite the widespread usage of ReLU, 
the research on how the activation functions may affect the flexibility of a network has not been well discussed. 

In this paper, 
we propose a novel activation function named Power Linear Unit (PoLU). 
We prove that networks using PoLU are able to increase the maximal number of response regions.
We further compare PoLU to other state-of-the-art activation functions in different networks and on different datasets. 
Experimental results demonstrate that PoLU outperforms almost all leading activation functions that have been widely used.

\begin{figure*}[t]
  \centering
  \includegraphics[width=15cm]{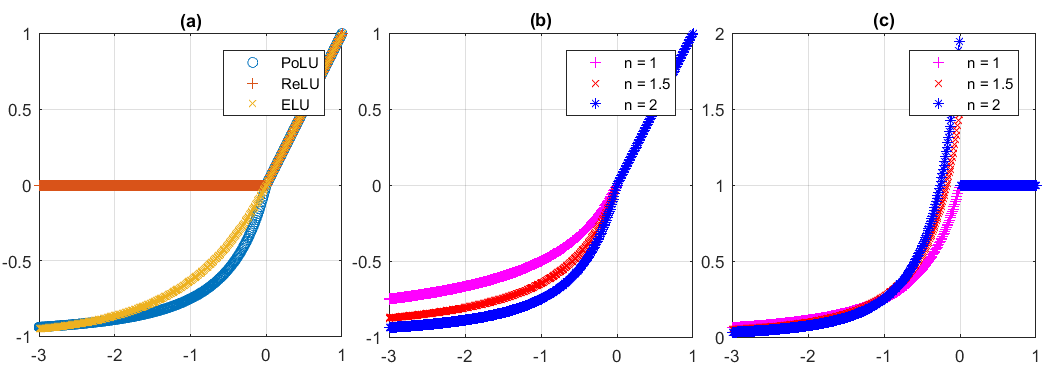}
  \caption{(a): the plot of PoLU ($n=2$), ReLU and ELU. (b): the plot of PoLU for $n = \{1, 1.5, 2\}$. (c): the plot of the derivatives of PoLU for $n = \{1, 1.5, 2\}$ }
  \label{fig:compare}
\end{figure*}

\section{Related Work}\label{sec:related}
Rectified Linear Unit (ReLU),
which is called ramp function sometimes,
was first applied to restricted Boltzmann machine (RBM) in \cite{relu_rbm}, and to neural networks \cite{relu} later.
It is among the most popular activation function nowadays.
The ReLU activation function can be defined as $f:\mathbb{R} \rightarrow \mathbb{R}$,
where $f(x) = \max(0,x)$,
and $x$ is the input of the neuron.
Comparing to traditional activation functions like the logistic sigmoid units or hyperbolic tangent units,
which are anti-symmetric,
ReLU is one-sided. 
This property encourages the network to be sparse ($i.e.$ the outputs of the hidden units are sparse),
and thus more biologically plausible.
In the experiments stated in \cite{relu},
the sparsity of the network can be from 50\% to 80\%.
Using ReLUs as activation functions also decreases the computational cost,
as they can be implemented via IF statements.
Most importantly,
ReLU alleviates the problem of vanishing gradient \cite{gradvan},
as the derivative of its positive part is always $1$.
This vanishing problem occurs in sigmoid and hyperbolic tangent units,
where the gradients vanish to $0$ after some epochs of training (due to the horizontal asymptotes) and stop the learning in the corresponding parts of the network.

Although ReLU has several advantages,
there are also some potential problems.
For instance,
ReLUs can "die" sometimes: once a neuron outputs $0$,
the corresponding weights may not be updated again,
since the gradients are also $0$.
Another issue is that,
since ReLU is non-negative,
the mean of the outputs of ReLUs in a layer will be positive,
which leads to the bias shift effect \cite{elu},
and may decrease the speed of learning. 

To overcome these problems,
researchers proposed some variants of ReLU.
Leaky Rectified Linear Unit (LReLU) \cite{lrelu} sets the output to be directly proportional to the input with a very small proportional constant $\alpha$ ($i.e.$ $\alpha = 0.01$).
This is equivalent to $f(x) = \max(x, \alpha x)$.
Under this definition of activation functions,
the neurons won't "die" as the gradient is small but still non-zero.
Parametric Rectified Linear Unit (PReLU) \cite{prelu} makes $\alpha$ to be a learnable parameter instead of a fixed constant. 
Since PReLU needs extra space and time complexity to learn $\alpha$s,
Randomized Leaky Rectified Linear Unit (RReLU) \cite{rrelu} was proposed,
where $\alpha$ is a random number sampled from a uniform distribution.
Shifted Rectified Linear Unit (SReLU) and Exponential Linear Unit (ELU) \cite{elu} both try to make the mean activation towards $0$ such that the gradients are closer to the natural gradient \cite{natgrad},
and hence speeding up the learning process. 
ELU also makes the gradient of its negative part to be non-zero to avoid dead neuron. 
However,
considering the function of ELU in the negative part,
$\alpha (e^x - 1) $,
we can see that $\alpha$ represents the slope of the function at $x \rightarrow 0^-$,
and the saturation value ($i.e.$ the value of $y$ when $x \rightarrow -\infty$) will be changed simultaneously if we vary the value of $\alpha$.
So, it is impossible to change the slope of the curve around $x = 0$ while keeping the asymptote $y = -1$.
In the latter section,
we demonstrate that under different saturation value,
the performance of $y = -1$ is a better choice for ELU (see Fig.\ref{fig:elu_alpha}). 
Therefore, we would like to develop an activation function where the slope at $x \rightarrow 0^-$ is independent of the asymptote of saturation.

For our proposed Power Linear Units (PoLUs),
we adopt the advantages of the activation functions mentioned above.
First, the output of PoLUs for positive input is designed to be identity to avoid gradient vanishing problem.
Second, PoLUs have non-zero output for negative inputs,
such that the output mean of the units is close to zero,
and thus reduce the bias shift effect.
Thirdly, there is a saturation on the negative part of PoLU,
which makes it more noise-robust for negative inputs.
Last but not least, PoLUs are able to map more portions of every layer's input space to the same range by using power function and thus increase the response regions of the neural network.
More details are to presented in the next section.


\section{Power Linear Unit}
In this section we propose PoLU and analyze the relation between the PoLU and the number of response regions.
Based on \cite{numreg2}, we redefine some terms such that the idea can be extended to a more general case. We first start by giving the definitions.

\begin{defn}
Let $f_n:\mathbb{R} \rightarrow \mathbb{R}$ be a function, the Power Linear Unit (PoLU) can be defined as follows:
\begin{equation}
    f_n(x) =
    \begin{cases}
        x                       &\text{if} \quad x \geq 0\\
        (1-x)^{-n} - 1  \quad   &\text{if} \quad x < 0
    \end{cases}
\end{equation}
and its derivative can be expressed as:
\begin{equation}
    f'_n(x) =
    \begin{cases}
        1                       &\text{if} \quad x \geq 0\\
        n(1-x)^{-n-1}   \quad   &\text{if} \quad x < 0
    \end{cases}
\end{equation}
\end{defn}
The parameter $n$ controls the rate of change of PoLUs at the negative part. Fig.\ref{fig:compare}(b) and (c) show the plots of PoLU and the derivative of PoLU under different power values $n$.
Similar to many previously proposed models of activation functions, keeping the identity as the positive section helps PoLUs against the gradient vanishing problem.
PoLUs also have non-zero outputs and a saturation plateau for negative inputs. These not only increase their ability to learn a stable representation, but also make the mean of the output of the units closer to zero, which reduces the bias shift effect \cite{elu}. 
In contrast to the previously proposed activation functions, PoLU has an intersection to $y = x$ at its negative regime when $n > 1$, which is proven to be good for increasing the number of response regions (see the proof of Theorem \ref{thm}).
Although ELU also has the same property when $\alpha > 1$, using such value of $\alpha$ may also push the mean activation away from zero (as $\alpha$ is a scaling factor), which typically leads to worse performance.

\begin{defn}
\label{def:i}
\cite{numreg2}
Let $F:\mathbb{R}^M \rightarrow \mathbb{R}^N$ be a map, $S \subseteq \mathbb{R}^M$ and $T \subseteq \mathbb{R}^M$.
$F$ identifies $S$ and $T$ if $F(S) = F(T)$.
\end{defn}

\begin{defn}
\label{def:rr}
A response region of a function $F$ is a maximal connected subset of the domain on which $F$ is differentiable and monotonic.
\end{defn}
As our proposed activation function is nonlinear for negative input,
we define response region in a more general way ($i.e.$ requiring $F$ to be differentialbe and monotonic instead of being just linear).
Under this definition,
Lemma 1 in \cite{numreg1} leads to Theorem \ref{thm:1layer},
and the proofs in \cite{numreg1,numreg2} still hold under definition \ref{def:rr}.
\begin{thm}
\label{thm:1layer}
\cite{numreg1}
The maximal number of response regions of the functions computed by a neural network, which has $n_0$ input units, one hidden layer with $n_1$ PoLUs,
is bounded below by $\sum_{j=0}^{n_0} \binom{n_1}{j}$.
\end{thm}

\begin{thm}
\label{thm}
The maximal number of response regions of the functions computed by a neural network,
which has $n_0$ input units,
$L$ hidden layers with $n_i \geq n_0$ PoLUs at the $i$-th layer,
is bounded below by
\begin{equation}
2^{n_0 (L-1)} \left ( \prod_{i=1} ^{L-1} \left \lfloor {\frac {n_i}{n_0}} \right \rfloor ^{n_0} \right )
\sum_{j=0}^{n_0} \binom{n_L}{j}
\end{equation}
\end{thm}
\begin{figure}[t]
  \centering
  \includegraphics[width=1\linewidth]{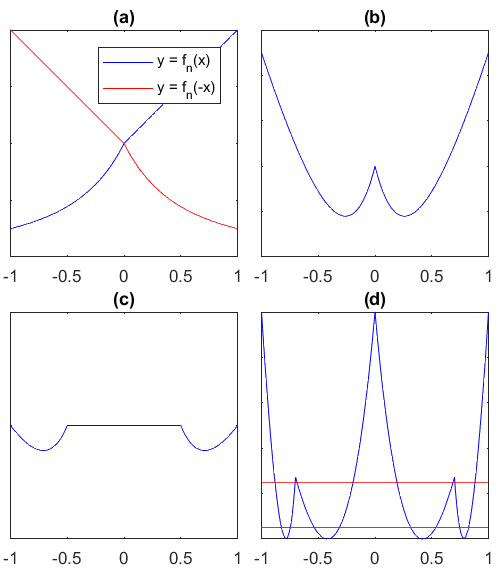}
  \caption{
    The plots of different curves for $n = 2$.
    (a) blue: $y = f_n(x)$; red: $y = f_n(-x)$;
    (b) $y = \hat{\varphi}_n (x)$;
    (c) $y = \varphi_n(x,0.5)$. The distance from the troughts to the origin is $0.5$;
    (d) blue: $y = S_2(x)$; red(upper): $y = b$; red(lower): $y = a$.
    The intervals which are subsets of $\{x \mid a < S_2(x) < b \}$ are mapped to the same set.
  }
  \label{fig:proof}
\end{figure}

\begin{proof}
We first start with two PoLUs.
Let $n > 1$ and $\hat{\varphi}_n : \mathbb{R} \rightarrow \mathbb{R}$ be a function defined as:
\begin{equation}
    \hat{\varphi}_n (x) = f_n(x) + f_n(-x)
\end{equation}
It is easy to prove that there is an intersection to $y = x$ at the negative region of PoLU for $n>1$, 
which causes two local minima exist in $y = \hat{\varphi}_n(x)$.
While a function formed by two ReLUs, following the construction in \cite{numreg2},
can only identify two regions, 
$\hat{\varphi}_n$,
which is formed by two PoLUs,
can identify four regions.

As the inputs always go through affine maps before they reach the activation functions,
we choose the weights and bias of these affine transformations.
Consider the modified version of $\hat{\varphi}_n$,
$\varphi_n : \mathbb{R} \times [0,1) \rightarrow \mathbb{R}$,
which is defined as follows:
\begin{equation}
    \varphi_n (x, d) = f_n(a_n(d)x+b_n(d)) + f_n(-a_n(d)x + b_n(d))
\end{equation}
We can always choose some suitable $a>0$ and $b>0$ to
(i) rescale $\varphi$ such that $\varphi(-1,d) = \varphi(0,d) = \varphi(1,d)$.
(ii) separate the two  troughs by a horizontal line with its length equals to $2d<1$,
Fig. \ref{fig:proof}(c) is the plot of $\varphi_n(x,0.5)$,
the distance from the troughs to the origin is equal to $0.5$.

\begin{figure*}[t]
    \centering
    \includegraphics[width=16cm]{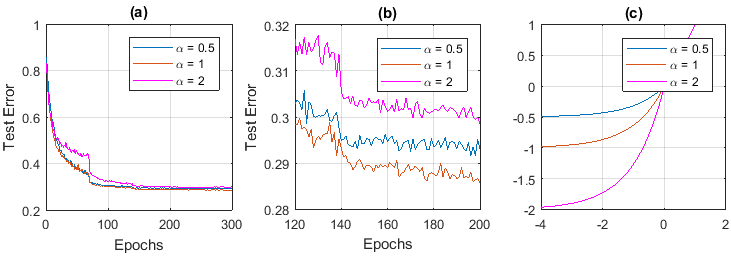}
    \caption{
        The plots show the results of using simple-ELU-Net on the CIFAR-100 dataset, by using ELU with $\alpha \in \{0.5, 1, 2\}$.
        (a-b) The testing error
        (c) The plots of ELU under different $\alpha$.
        ELU with $\alpha = 1$ has the best performance among them.
    }
  \label{fig:elu_alpha}
\end{figure*}

Consider a layer of $n_1$ PoLUs with $n_0$ input, where $n_1 \geq n_0$.
The PoLUs are separated into $n_0$ disjoint subsets of cardinality $p$, which is the largest even number not greater than $n_1/n_0$ ($i.e.$ $p = 2k \leq \lfloor n_1/n_0 \rfloor$ for some  $k \in \mathbb{N}$), the remaining PoLUs are ignored.

Without loss of generality, we consider the $j$-th input of the layer, where $j \in \{ 1, ..., n_0 \}$.
As we can choose the input weights and biases, the functions can be constructed in the following way:
\begin{equation} \label{eq:h}
    \begin{split}
    h_1(x) &= \varphi_n(x, d_1)  \\
    h_2(x) &= \varphi_n(x, d_2)   \\
            &\vdots                  \\
    h_k(x) &= \varphi_n(x, d_k)
    \end{split}
\end{equation}
and let $(-c_i, h_i(-c_i))$ and $(c_i, h_i(c_i))$ be the coordinates of the local minima of $h_i(x)$ for $i \in \{1,2,...,k\}$.

We use a pair of $f_n$ to construct one function. Fig.\ref{fig:proof}(a) and (b) show the plots of $h_1$ and $h_2$ respectively.

We can construct a function defined as follows:
\begin{align}
    S_N(x) = &\sum_{i=1}^N a_ih_i(x) \quad     \forall N \in \{1, ..., k\} \\
    &\tilde{h}(x) = S_k(x)
\end{align}
For $h_1$, we set $d_1 = 0$ such that the two  troughs are stick together.
For $h_2$, we set $d_2 > c_1$ to ensure that there are four local minima for the function $y = S_2(x) = a_1 h_1(x) + a_2h_2(x)$ for some constant $a_1$, $a_2$. 
We keep constructing $S_i$ with $i>2$ in a similar way.
In general, we have $d_1 = 0 < c_1 < d_2 < c_2 < d_3 < c_3 < ... < d_k < c_k < 1$.
Under this setting, we have $2k$  troughs for $y = S_k(x)$.

By choosing suitable coefficients $\{a_i\}$,
we can construct a function with $p = 2k$ local minima,
where they have a same value at $y$-coordinates.
Fig.\ref{fig:proof}(d) shows the plot of $y = S_2(x)$,
for the 8 intervals 
$[-1,-c_2]$,
$[-c_2,-d_2]$,
$[-d_2,-c_1]$,
$[-c_1,0]$,
$[0,c_1]$,
$[c_1, d_2]$,
$[d_2,c_2]$,
$[c_2,1]$,
there exists at least one subset of each interval such that these subsets are mapped onto the same interval by $S_2$.

Now, for the following intervals:
\begin{equation}
\begin{split}
    &[-1, -c_k],
    [-c_k, -d_k],
    [-d_k,-c_{k-1}],...,    \\
    &[-c_1,0],
    [0,c_1],
    [c_1,d_2],...,
    [c_{k-1}, d_k],
    [d_k,c_k],
    [c_k,1]
\end{split}
\end{equation}
there is a subset of each intervals that is mapped onto the same interval.
Therefore, $\tilde{h}$ identifies $2p = 4k$ regions in the input domain. 

Since $\tilde{h}$ is the linear combination of $h_1$, ..., $h_k$, we can treat it as an output of the current layer of the neural network.
By considering all the $n_0$ subset,
we can conclude that,
$\tilde{h}$ identifies $2^{n_0} \lfloor n_1/n_0 \rfloor ^{n_0}$ regions in total.

Using the same strategy as \cite{numreg1,numreg2},
we can conclude that,
the maximal number of response regions computed by the neural network using PoLUs is bounded below by
$2^{n_0 (L-1)} \left ( \prod_{i=1} ^{L-1} \left \lfloor {\frac {n_i}{n_0}} \right \rfloor ^{n_0} \right ) \sum_{j=0}^{n_0} \binom{n_L}{j}$.
\end{proof}
This proof holds for the neural networks with PoLUs($n > 1$) or ELUs($\alpha > 1$),
as they have intersection with $y=x$ for negative input.
While for the networks with PoLUs($n \leq 1$), ELUs($\alpha \leq 1$) or ReLUs,
the lower bounds remain the same as in \cite{numreg2}.
As the number of response regions of the functions computed by a neural network, is a measure of flexibility of that network,
the networks with PoLUs($n>1$) and ELUs($\alpha>1$) are considered to be more flexible.

Note that, even though we prove that the PoLU has more response regions than ELU does, we cannot simply determine that any new activation functions which may increase the number of response regions can have a better performance. Because there are many factors that may affect the performance of a network, like the bias shift effect, value of the saturation for negative input, etc. Hence, "the larger the number of response regions, the better the network is" is not the final conclusion and not the only advantage that PoLU has.


\begin{table*}[]
\centering
\caption{This table shows the result for MNIST dataset and SVHN dataset. For SVHN, Simple-ELU-Net and a 4CNN+2NN network are used for testing. For MNIST, only a 2CNN+2NN is used for testing. We use ReLU, ELU and PoLU($n = 1$, $1.5$, $2$) for both datasets. The best result for each network is bold.}
\begin{tabular}{lllll}
\hline
              & \multicolumn{1}{|c|}{\multirow{2}{*}{MNIST}} & \multicolumn{2}{c}{SVHN}               \\ \cline{3-4}
              & \multicolumn{1}{|c|}{}                       & \multicolumn{1}{l|}{simple-ELU-Net} & 4 CNN + 2 NN   \\
\hline
ReLUs
& \bm{$0.69(\pm0.03)$}\%
& $4.85(\pm0.07)$\%
& $5.18(\pm0.09)$\% \\
ELUs
& $0.99(\pm0.02)$\%
& $4.87(\pm0.05)$\%
& $5.16(\pm0.08)$\% \\
PoLUs (n=1)
& $0.83(\pm0.03)$\% 
& $4.84(\pm0.06)$\%
& $5.02(\pm0.08)$\% \\
PoLUs (n=1.5)
& $0.83(\pm0.02)$\%
& $4.71(\pm0.05)$\%
& $4.96(\pm0.07)$\% \\
PoLUs (n=2)
& $0.87(\pm0.02)$\%
& \bm{$4.63(\pm0.06)$\%}
& \bm{$4.90(\pm0.07)$\%} \\
\hline
\end{tabular}
\label{tab:mnistsvhn}
\end{table*}

\section{Experiments} \label{sec:exp}
In this section,
we first evaluate the impact of $\alpha$ on ELUs by setting $\alpha$ to 0.5, 1
and 2 with the ELU-Network on CIFAR-100 dataset.
We then evaluate PoLUs with different convolutional neural networks using different power values $n \in \{1, 1.5, 2\}$ on five benchmark datasets:
MNIST \cite{mnist}, CIFAR-10 \cite{cifar},
CIFAR-100 \cite{cifar},
Street View House Numbers \cite{svhn},
and ImageNet \cite{imagenet}.
Compared with other state-of-the-art activation functions,
including Exponential Linear Units (ELUs),
and the most widely used activation function,
Rectified Linear Units (ReLUs),
convolutional neural networks with PoLUs present the best performance on all four datasets.
Several deep neural networks with a different number of layers are implemented to demonstrate that PoLUs are more compatible than ELUs and ReLUs for convolutional neural networks.
The experiments are implemented using deep learning toolbox Keras \cite{keras} with tensorflow backend.
Note that, we run experiments with activation functions SReLU and Leaky ReLU as well, obtained similar results comparable to those reported in the ELU paper ($i.e.$ the accuracy difference is within $1\%$). Therefore, to make the plots more clear and easier to demonstrate the differences between ELU, PoLU and the baseline activation function ReLU, we do not contain the performance of SReLu and Leaky ReLU in the plots.

\subsection{MNIST dataset}
The MNIST dataset \cite{mnist} contains 60,000 training and 10,000 testing samples with $28 \times 28$ pixel size.
Each image is drawn from greyscale handwritten digits 0-9.
Since the MNIST dataset has been well studied in evaluating different neural networks,
we utilize this dataset to assess the performance of PoLUs with a relatively shallow neural network.
We train a network with two convolutional layers and two densely connected layers followed by a softmax layer which is arranged in stack of ($[1 \times 32 \times 3], [1 \times 64 \times 3], [1 \times 128 \times FC], [1 \times 10 \times Softmax]$). A $2 \times 2$ max-pooling layer with stride of 2 is applied to the end of the second stack.
We leverage dropout with a ratio 0.5 to regularize the network. The results are provided in Table \ref{tab:mnistsvhn},and ReLUs achieved the best results with the testing error equaling to 0.69\%.

\begin{figure*}[t]
  \centering
  \includegraphics[width=16cm]{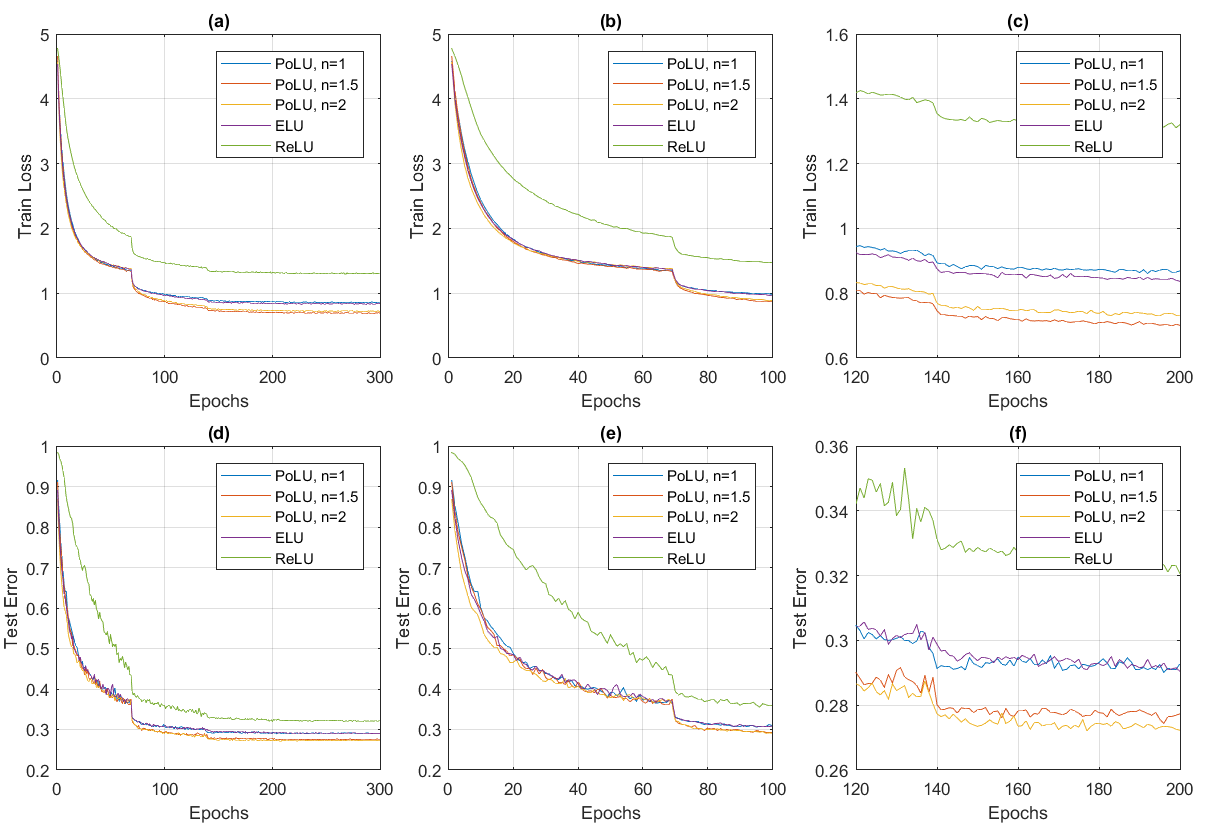}
  \caption{These plots are the results of using Simple-ELU-Net and different activation functions (PoLU($n\in\{1,1.5,2\}$), ELU, ReLU) on CIFAR-100 dataset. 
        (a)-(c): The training loss;
        (d)-(f): The testing error.
        Each curve represents the average of 5 runs. PoLU with $n = 2$ achieves the best performance.
        }
  \label{fig:eval}
\end{figure*}

\subsection{CIFAR-10 and CIFAR-100 dataset}
The CIFAR-10 and CIFAR-100 \cite{cifar} datasets are similar. Both of these two datasets contain 60,000 color images with $32 \times 32$ size, which are split into 50,000 training and 10,000 testing samples. The only difference between them is that CIFAR-10 is drawn from 10 classes while CIFAR-100 contains 100 classes. Therefore, the same neural network structures followed by different softmax layer are implemented for both datasets. The neural networks we implemented are (i) the ELU-Network with 11 convolutional layers from \cite{elu}(named simple-ELU-Net), (ii) the ELU-Network with 18 convolutional layers from \cite{elu}(named ELU-Net), (iii) A VGG16-structure-like neural network \cite{vgg16}, and (iv) Deep Residual Network (ResNet) with 50-layer structure from \cite{resnet}. As mentioned before, we also implement the ELU-Network with 11 convolutional layers with ELUs assigned with different $\alpha$ values to evaluate the effect of $\alpha$ and the relationship between slope and saturation. 

The assessment of PoLUs assigned with different power values ${n}$ are mainly based on the comparison with ELUs and ReLUs by utilizing the ELU-Network containing 11 convolutional layers on CIFAR-100 dataset. The relatively simple ELU-Network is arranged in stacks of ($[1 \times 192 \times 5], [1 \times 192 \times 1, 1 \times 240 \times 3], [1 \times 240 \times 1, 1 \times 260 \times 2], [1 \times 260 \times 1, 1 \times 280 \times 2], [1 \times 280 \times 1, 1 \times 300 \times 2], [1 \times 300 \times 1], [1 \times 100 \times 1]$) (in each stack, the first number is the number of layers, the second number represents the number of filters and the third number is the size of each filter). After each stack (except the last two stacks), a $2 \times 2$ max-pooling layer with stride of 2 is applied. Followed by the instruction in \cite{elu}, the dropout ratio of ($0.0, 0.1, 0.2, 0.3, 0.4, 0.5, 0.0$) is applied at the last layer of each stack and the L2-weight decay regularization is set to 0.0005. Learning rate is initialized at 0.01 and after 70 epochs drop to 0.005. Then learning rate decays by a factor of 10 every 70 epochs. The total number of epochs is 300. Stochastic Gradient Descent is used with momentum set to 0.9. Global contrast normalization and ZCA whitening which are mentioned in \cite{maxout} are also applied to the dataset. Moreover, the training samples are randomly cropped into $32 \times 32$ size with random horizontal flipping images padded with four pixels at all borders. Each network with different activation function is trained 5 times, and the mean and standard deviation of the testing errors on CIFAR-10 and CIFAR-100 are shown in Table \ref{tab:cifar10} and \ref{tab:cifar100}. Fig.\ref{fig:eval} presents the comparison of different activation functions based on the testing error and training loss on CIFAR-100. The results demonstrate that the PoLUs with power value $n > 1$ achieves the best results among the others, which also satisfy our prediction. Additionally, training loss of PoLUs with different power values drops faster than ReLU and with a comparative speed with ELU.

\begin{table*}[t]
\centering
\caption{This table shows the result for CIFAR-10 dataset. Different Networks (simple-ELU-Net, ELU-Net, VGG16, ResNet-50) and activation function (ReLU, ELU, PoLU($n = 1$, $1.5$, $2$)) are used. The best result for each network is bold.}
\begin{tabular}{lllll}
\hline
              & simple-ELU-Net & ELU-Net & VGG16 & ResNet-50 \\
\hline
ReLUs
& $11.73(\pm0.20)$\%
& $8.37(\pm0.19)$\%
& $8.16(\pm0.11)$\%
& $9.44(\pm0.20)$\% \\
ELUs
& $9.92(\pm0.13)$\%
& $6.67(\pm0.11)$\%
& $7.46(\pm0.13)$\%
& $8.76(\pm0.17)$\%  \\
PoLUs (n=1)
& $9.87(\pm0.11)$\%
& $6.68(\pm0.10)$\%
& $7.57(\pm0.10)$\%
& $8.64(\pm0.13)$\%  \\
PoLUs (n=1.5)
& $9.03(\pm0.12)$\%
& $5.85(\pm0.15)$\%
& \bm{$7.03(\pm0.14)$\%}
& \bm{$8.14(\pm0.13)$\%}  \\
PoLUs (n=2)
& \bm{$8.74(\pm0.11)$\%}
& \bm{$5.45(\pm0.10)$\%}
& $7.35(\pm0.12)$\%
& $8.42(\pm0.14)$\%  \\
\hline
\end{tabular}

\label{tab:cifar10}
\end{table*}


To evaluate deeper and more complex neural networks which contain more parameters, we implemented both a VGG16-structure-like neural network and the ELU-Network with 18 convolutional layers. The VGG16-structure-like neural work is derived from \cite{vgg16}, with the number of units in the last two densely connected layers changed to 512 since the size of the input feature maps of them are $1 \times 1 \times 512$. The structure of the neural network is arranged in stacks of ($[2 \times 64 \times 3], [2 \times 128 \times 3], [3 \times 256 \times 3], [3 \times 512 \times 3], [3 \times 512 \times 3], [2 \times 512 \times FC], [1 \times 100 \times FC]$). Max-pooling with size $2 \times 2$ and stride 2 is applied after each stack except the last three densely connected layers. The dropout ratio is set to 0.5 for the last two densely connected layers and the L2-regularization term for each convolutional layer is set to 0.0005. The learning rate is initialized to 0.01 and decays by a factor of 10 every 70 epochs. The optimizer for the neural network is SGD with momentum equals to 0.9.
The more sophisticated ELU-Network is arranged in stacks of ($[1 \times 384 \times 3], [1 \times 384 \times 1, 1 \times 384 \times 2, 1 \times 640 \times 2], [1 \times 640 \times 1, 3 \times 768 \times 2], [1 \times 768 \times 1, 2 \times 896 \times 2], [1 \times 896 \times 3, 2 \times 1024 \times 2], [1 \times 1024 \times 1, 1 \times 1152 \times 2], [1 \times 1152 \times 1], [1 \times 100 \times 1]$). Similar to the simpler ELU-Network mentioned above, there is a $2 \times 2$ max-pooling layer with stride 2 after each stack except the last two stacks. Zero padding is applied before each convolutional layer to keep the dimension unchanged. The initial dropout ratio, L2-Regularization, momentum value are the same as the simpler ELU-Network. For both the VGG16-structure-like network and the sophisticated ELU-Network, the dataset is preprocessed as described in \cite{maxout} with global contrast normalization and ZCA whitening.

Thanks to the success of much deeper neural networks on image classification, the experiments with different activation functions should also take deeper neural network such as Deep Residual Network (ResNet) into consideration. Comparing with the neural networks mentioned above, ResNet is much deeper and the structure is kind of different since there is a shortcut connection from each stack input to the next stack input. As described in \cite{resnet}, this skip connection structure can efficiently solve the vanishing gradient problem which is described in a previous section.
Another difference is, there is always a batch normalization layer before each activation function. Batch normalization can provide the input batch of each stack with zero mean and unit variance, which will compensate one of the disadvantages of ReLUs. 
Therefore, the difference of the performance among ReLUs, ELUs, and PoLUs in ResNet is not as significant, compared with other networks. Such phenomenon is also observed in Table \ref{tab:cifar10} and \ref{tab:cifar100}.
However, we implemented ResNet-50 with the last two residual stages ending up by utilizing a $1 \times 1$ convolutional layer with stride of 2 to downsample the feature maps, thus it will result in a worse performance than what is described in \cite{resnet}. The difference of the test errors of ResNet-50 with ReLUs on CIFAR-10 dataset is about $3\%$.

The evaluation of different $\alpha$ of ELUs on CIFAR-100 dataset using ELU-Network with 11 convolutional layers is presented in Fig.\ref{fig:elu_alpha}.
The results show that ELUs with $\alpha= 1$ could achieve a better performance compared with  $\alpha = 0.5$ or $\alpha = 2$, since the saturation value is too small or too large, which may push the mean activation away from 0. This is also the evidence of the disadvantages of ELUs, which we mentioned in a previous section.

\begin{table*}[t]
\centering
\caption{This table shows the result for CIFAR-100 dataset. Different Networks (simple-ELU-Net, ELU-Net, VGG16, ResNet-50) and activation function (ReLU, ELU, PoLU($n = 1$, $1.5$, $2$)) are used. The best result for each network is bold.}
\begin{tabular}{lllll}
\hline
              & simple-ELU-Net & ELU-Net & VGG16 & ResNet-50 \\
\hline
ReLUs
& $33.53(\pm0.41)$\%
& $29.47(\pm0.31)$\%
& $32.87(\pm0.21)$\%
& $30.27(\pm0.31)$\% \\
ELUs
& $29.06(\pm0.25)$\%
& $24.88(\pm0.21)$\%
& $29.29(\pm0.17)$\%
& $28.21(\pm0.29)$\% \\
PoLUs (n=1)
& $29.02(\pm0.34)$\%
& $24.97(\pm0.22)$\%
& $28.90(\pm0.22)$\%
& $28.46(\pm0.27)$\% \\
PoLUs (n=1.5)
& $27.89(\pm0.26)$\%
& $24.05(\pm0.23)$\%
& \bm{$27.48(\pm0.23)$}\%
& \bm{$28.01(\pm0.29)$}\% \\
PoLUs (n=2)
& \bm{$27.01(\pm0.21)$\%}
& \bm{$23.07(\pm0.20)$\%}
& $28.64(\pm0.21)$\%
& $28.13(\pm0.30)$\% \\
\hline
\end{tabular}

\label{tab:cifar100}
\end{table*}

\subsection{Street View House Number (SVHN) dataset}
The Street View House Number (SVHN) \cite{svhn} dataset is collected by Google Street View, focusing on the color images of house numbers. There are two formats of this dataset, and the second one was what used in our experiments. In the second format, each image is in the fixed size with $32 \times 32$ pixels and most center part of the image is a digit. There are 73,257 training and 26,032 testing samples. Additionally, there are also 531,131 extra samples that can be used as additional training samples. SVHN can be viewed as similar to MNIST since both datasets are mainly focused on digits. However, SVHN is harder than MNIST since images from SVHN are cropped from real world color images. Moreover, the size of each image is $32 a\times 32$, and thus neural networks which are designed for CIFAR-10 and CIFAR-100 can also be implemented for SVHN. We implemented the ELU-Network with 11 convolutional layers and a relatively shallower neural network which contains four convolutional layers and one densely connected layer followed by a softmax layer. The relatively shallow neural network is arranged in stack of ($[1 \times 32 \times 3], [1 \times 32 \times 3], [1 \times 64 \times 3], [1 \times 64 \times 3]. [1 \times 512 \times FC], [1 \times 10 \times Softmax]$). The dataset is preprocessed followed by \cite{svhnprepreocess} with local contrast normalization. Only training and testing samples are used, the set of extra samples is not considered. The final performance is shown in Table \ref{tab:mnistsvhn}.

\begin{figure}[t]
  \centering
  \includegraphics[width=1\linewidth]{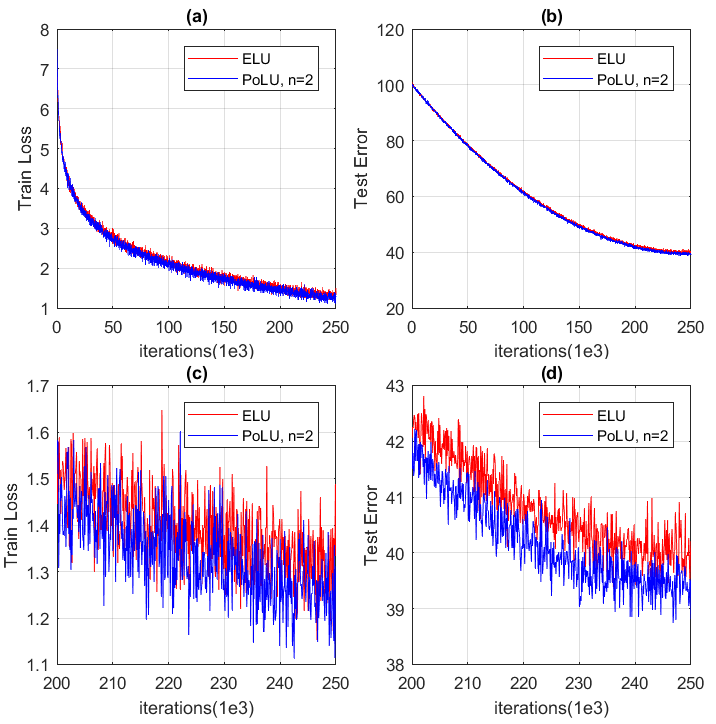}
  \caption{The plots of the training loss and testing error(\%) on ImageNet.
  (a) The training loss on the 1 - 250k$th$ iterations;
  (b) The testing error(\%) on the 1 - 250k$th$ iterations;
  (c) The training loss on the 200k -250k$th$ iterations;
  (d) The testing error(\%) on the 200k - 250k$th$ iterations.
  }
  \label{fig:imagenet}
\end{figure}

\subsection{ImageNet}
We evaluate PoLU on the ImageNet dataset which contains more than 1.2 million training images belonging to 1000 classes. There are also 50k and 100k images for validation and testing respectively. The network we implemented has the same structure as the one proposed in \cite{elu}, in which the 15 layers are now arranged as ($[1 \times 96 \times 6], [3 \times 512 \times 3], [5 \times 768 \times 3], [3 \times 1024 \times 3]. [2 \times 4096 \times FC], [1 \times 1000 \times Softmax]$).
After each stack, a $2 \times 2$ max-pooling layer with stride of 2 is applied and spatial pyramid pooling (SPP) with [1,2,4] levels is employed before the first FC layer \cite{prelu}.  

In Fig.\ref{fig:imagenet}, even though the network with ELU has similar performance with network with PoLU in the $(a)$ and $(b)$ plots , we can still observe that, comparing with the network with ELUs,
the one with PoLUs not only has a slightly lower loss at the last several epochs,
but also has a better final accuracy in plots $(c)$ and $(d)$.

\subsection{Discussion on the result}
The performance of different neural networks with different activation functions illustrates that the PoLUs and ELUs are much better than ReLUs in deep neural network if there are no batch normalization layers, which indicates that PoLUs and ELUs overcome the bias shift problem and push the input's mean to zero from the other side. However, the result on MNIST dataset shows that ReLU work better than ELU and PoLU with $n = 1,1.5,2$, which suggests that the bias shift effect may have less influence in shallow neural networks.
Another observation from the comparison of the ELU-Network and the VGG16-structure-like network on CIFAR-100 dataset is that, if the network consists of all convolutional layers, like the ELU-Network, the PoLUs and ELUs, which contain negative part, it will perform much better than ReLUs, while if there are densely connected layers, the difference in performance between PoLUs, ELUs and ReLUs are not so significant.
In our experiments, we also implemented power value $n > 2 $. However, due to too-large a slope at $x\rightarrow 0^-$, the activation functions are over sensitive to the input values. PoLUs will map inputs around $0^-$ with a little difference into larger area and PoLUs also reach to saturation faster as the rate of change towards saturation is increased. Moreover, together with the experiments we implemented with ELUs by setting different values of $\alpha$, we can draw the conclusion that both slope and saturation cannot be set too large.

For the time complexity of our experiments, as demonstrated in both Fig. \ref{fig:eval} and Fig. \ref{fig:imagenet}, similar to ELU, PoLU can achieve the same training loss or training accuracy with less epochs compared with ReLU. Therefore, even though PoLU is more complex than ReLU and it takes about $5\%$ more training time for each training epoch, the time of reaching the best accuracy that ReLU can get is still less. Note that the extra time spent on the calculation of activation function is only a small portion, comparing to the total computational time of the network. While training with same number of epochs, the total training time for ELU/PoLU network and ReLU network on ImageNet are 12 hours and 11.5 hours respectively.

\section{Conclusions}
We proposed a new activation function - Power Linear Unit (PoLU),  which uses identity function and power function to construct its positive and negative sections respectively. Even though the PoLU is along the direction of the efforts like ELU, which are variants of ReLU, in PoLU more attention is on the negative part of the activation function. This was motivated by the observation that not only the saturation value but also the slope in the negative part of the activation function can significantly affect the performance of the network. In contrast, ELU focuses on better saturation values via using exponential computation. Note that the saturation value will change if we change the slope. We have shown that ELU with saturation value $1$ performs best, which implies that the slope cannot be changed if we want to use this “optimal” value. The proposed PoLU can avoid this dilemma: we can achieve the desired saturation while being able to adjust the slope based on changing the power parameter $n$, which means that the negative region of PoLU can intersect $y = x$ while keeping the same asymptote for the saturation by means of setting $n > 1$. We also demonstrated the advantage of having a proper slope in the negative part of the activation function, that networks using PoLU may have a larger number of response regions, which helps to improve the nonlinearity capacity of the neural network.
Experimental results showed that PoLU outperforms other state-of-the-art on most networks.

{\small
\bibliographystyle{ieee}
\bibliography{egbib}
}

\end{document}